\documentclass{article}

\usepackage{graphicx,amscd,amsmath,amssymb,verbatim}
\usepackage[dvips]{hyperref}
\usepackage[TS1,OT1,T1]{fontenc}

\begin{document}

\title{Using Self-Organising Mappings to Learn the Structure of
Data Manifolds}
\author{Stephen Luttrell}
\maketitle

\noindent {\bfseries Abstract:} In this paper it is shown how to
map a data manifold into a simpler form by progressively discarding
small degrees of freedom. This is the key to self-organising data
fusion, where the raw data is embedded in a very high-dimensional
space (e.g. the pixel values of one or more images), and the requirement
is to isolate the important degrees of freedom which lie on a low-dimensional
manifold. A useful advantage of the approach used in this paper
is that the computations are arranged as a feed-forward processing
chain, where all the details of the processing in each stage of
the chain are learnt by self-organisation. This approach is demonstrated
using hierarchically correlated data, which causes the processing
chain to split the data into separate processing channels, and then
to progressively merge these channels wherever they are correlated
with each other. This is the key to self-organising data fusion.
\section{Introduction}

The aim of this paper is to illustrate an approach that maps raw
data into a representation that reveals its internal structure.
The raw data is a high-dimensional vector of sample values output
by a sensor such as the samples of a time series or the pixel values
of an image, and the representation is typically a lower dimensional
vector that retains some or all of the information content of the
raw data. There are many ways of achieving this type of data reduction,
and this paper will focus on methods that learn from examples of
the raw data alone.

A key approach to data reduction is the self-organising map (SOM)
\cite{Kohonen2001}. There are many variants of the SOM approach
which may be used to map raw data into a lower dimensional space
that retains some or all of its information content. In order to
increase the variety of mappings SOMs can learn some of these variants
use quite sophisticated learning algorithms. For instance, the\ \ topology
of a SOM can be learnt by the neural gas approach \cite{MartinezSchulten1991},
or the topology of the network connecting several SOMs can be learnt
by the growing hierarchical self-organising map (GHSOM) approach
\cite{DittenbachMerklRauber2000}.

The approach used in this paper aims to achieve a similar type of
result to the GHSOM approach. GHSOM is a top-down coarse-to-fine
approach to optimising a tree structured network of SOMs, whereas
in this paper a bottom-up fine-to-coarse approach will be used that
learns a tree structure where appropriate. The choice of a fine-to-coarse
rather than coarse-to-fine approach is made in order to obtain networks
that can be readily applied to data fusion problems, where the goal
is to progressively discard noise (and irrelevant degrees of freedom)
as the data passes along the processing chain, thus gradually reducing
its dimensionality to eventually obtain a low-dimensional representation
of the original raw data.

The basis for the approach used in this paper is a Bayesian theory
of SOMs \cite{Luttrell1994} in which a SOM is modelled as an encoder/decoder
pair, where the decoder is the Bayes inverse of the encoder. In
this approach the encoder is modelled as a conditional probability
over all possible codes given the input, and the code that is actually
used is a {\itshape single} sample drawn from this conditional probability
(i.e. a winner-take-all code). When the conditional probability
is optimised to minimise the average Euclidean distortion between
the original input and its reconstruction this leads to a network
that has properties very similar to a Kohonen SOM.

The basic approach \cite{Luttrell1994} needs to be extended in two
separate ways \cite{Luttrell2003}. Firstly, to encourage the self-organisation
of a processing chain leading from raw data to a higher level representation,
the single encoder/decoder is extended to become a Markov chain
of connected encoders, where each encoder feeds its output into
the next encoder in the chain. Secondly, to encourage the self-organisation
of each encoder into a number of separate smaller encoders and thus
to learn tree-structured networks where appropriate, each encoder
is generalised to use codes that make simultaneous use of {\itshape
several} samples from the conditional probability rather than only
a {\itshape single} winner-take-all sample.

The goal of the approach used in this paper is similar to that of
the multiple cause vector quantisation approach \cite{RossZemel2003},
because the common aim is to split data into its separate components
(or causes). However, the approach used in this paper aims to minimise
the amount of manual intervention in the training of the network,
and thus allow the structure of the data to determine the structure
of the network. This is made possible by using codes that consist
of {\itshape several} samples from a {\itshape single} conditional
probability, which allows each encoder to decide for itself how
to split into a number of separate smaller encoders. Also the approach
used in this paper does {\itshape not} make explicit use of a generative
model of the data, because the aim is only to map raw data into
a representation that clarifies its internal structure (i.e. build
a recognition model), for which a generative model may be sufficient
but is actually not necessary.

This paper is organised as follows. In Section \ref{XRef-Section-82210171}
the structure of data is represented as smooth curved manifolds,
and encoders are represented as hyperplanes that slice through these
manifolds. In Section \ref{XRef-Section-822101712} the theory of
a single encoder is developed by extending a Bayesian theory of
SOMs \cite{Luttrell1994} from winner-take-all encoders to multiple
output encoders, and this theory is further extended to Markov chains
of connected encoders \cite{Luttrell2003}. In Section \ref{XRef-Section-822101724}
these results are used to train a network on some hierarchically
correlated data to demonstrate the self-organisation of a tree-structured
network for processing the data.
\section{Data Manifolds}\label{XRef-Section-82210171}

In order to represent the structure of data a flexible framework
needs to be used. In this paper an approach will be used in which
the structure of the data manifold is of primary importance, and
the aim is to split apart the manifold in such a way as to reveal
how its overall structure is composed. This approach must take account
of the relative amplitude of the various contributions, so that
a high resolution representation would include even the smallest
amplitude contributions to the manifold, and a low resolution representation
would retain only the largest amplitude contributions. More generally,
it would be useful to construct a sequence of representations, each
with a lower resolution than the previous one in the sequence. This
could be achieved by progressively discarding the smallest degree
of freedom to gradually lower the resolution of the representation.
In effect, the representation will become increasingly abstract
as it becomes more and more invariant to the fine details of the
original data manifold.

In Section \ref{XRef-Section-822101816} the basic notation used
to describe manifolds is presented, and in Section \ref{XRef-Section-822101823}
the process of splitting a manifold into its component pieces and
then reassembling these to form an approximation to the manifold
is described.
\subsection{Representation of Data Manifolds}\label{XRef-Section-822101816}

Assume that the raw data vector $x$ lies on a smooth manifold $x(
u) $, parameterised by $u$ which is a vector of co-ordinates in
the manifold. Usually, though not invariably, $x$ is a high-dimensional
vector (e.g. an image comprising an array of pixel values) and $u$
is a low-dimensional vector (e.g. a vector of object positions),
in which case the space in which $x$ lives is a high-dimensional
embedding space for a low-dimensional manifold. Typically, $u$ represents
the underlying degrees of freedom (e.g. object co-ordinates), whereas
$x$ represents the observed degrees of freedom (e.g. sensor measurements).
Usually, $u$ will contain some noise degrees of freedom, but these
can be handled in exactly the same way as other degrees of freedom
by splitting $u$ as $u=(u_{s},u_{n})$ where $u_{s}$ is signal and
$u_{n}$ is noise. The probability density function (PDF) $\Pr (
u) $ describes how the manifold is populated and $\Pr ( x) $ (where
$\Pr ( x) =\int du \Pr ( u) \delta ( x-x( u) ) $) describes how
the embedding space is populated.

In general, $x( u) $ is a non-linear function of $u$ so the manifold
is curved, and thus occupies more linear dimensions of the embedding
space than would be the case if the manifold were not curved. It
is commonplace for a 1-dimensional manifold (i.e. $u$ is a scalar)
to be curved so as to occupy {\itshape all} of the linear dimensions
of the embedding space (e.g. the manifold of images generated by
moving an object along a 1-dimensional line of positions).
\begin{figure}[h]
\begin{center}
\includegraphics{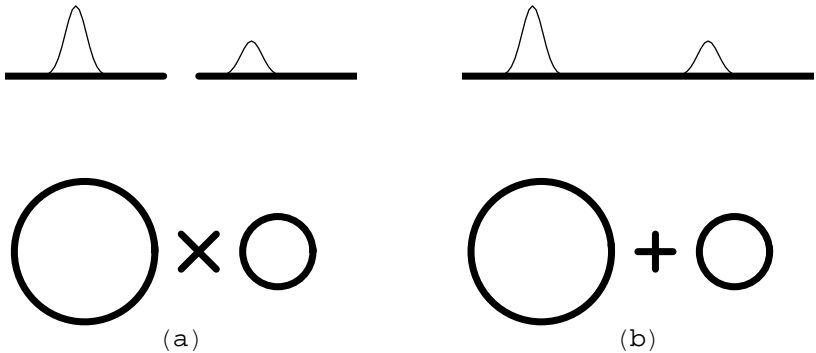}

\end{center}
\caption{Examples of manifolds generated by images of a pair of
objects. In each of the two diagrams the upper half shows the sensor
data, and the lower half shows the low-dimensional manifold {\itshape
topology} assuming that the sensor data have circular wraparound.
The high-dimensional manifold {\itshape geometry} has a number of
dimensions equal to the number of pixels in the corresponding sensor.
(a) Tensor product of manifolds: 2-torus topology. This is generated
by observing each object using a separate sensor.\ \ (b) Superposition
(or mixture) of manifolds: This is generated by observing both objects
using the same sensor, so that there is the possibility of overlap
(possibly with obscuration) of the sensor data from the two objects.
In the limit where the objects overlap infrequently case (b) closely
approximates case (a).}\label{XRef-Figure-822102129}
\end{figure}

If $x( u) $ can be written as $x( u) =(x_{1}( u_{1}) ,x_{2}( u_{2})
)$, where $x_{1}( u_{1}) $ and $x_{2}( u_{2}) $ are independently
parameterised manifolds living in separate subspaces of the embedding
space (where $\dim  x=\dim  x_{1}+\dim  x_{2}$ and $\dim  u=\dim
u_{1}+\dim  u_{2}$), then $x( u) $ describes a tensor product of
manifolds as shown in Figure \ref{XRef-Figure-822102129}a. This
type of manifold arises when the underlying degrees of freedom are
measured by separate sensors. This parameterisation can readily
be generalised to $x( u) =(x_{1}( u_{1}) ,x_{2}( u_{2}) ,\cdots
,x_{k}( u_{k}) )$ for $k>2$. For independently populated manifolds
$\Pr ( u) $ factorises as $\Pr ( u) =\Pr ( u_{1}) \Pr ( u_{2}) $,
and if $x( u) =(x_{1}( u_{1}) ,x_{2}( u_{2}) )$ then $\Pr ( x) =\Pr
( x_{1}) \Pr ( x_{2}) $ where $\Pr ( x_{i}) =\int du_{i} \Pr ( u_{i})
\delta ( x_{i}-x_{i}( u_{i}) ) $ for $i=1,2$. For manifolds that
are populated in a correlated way (i.e. $\Pr ( u) \neq \Pr ( u_{1})
\Pr ( u_{2}) $) no such simple result holds.

If $x( u) $ can be written as $x( u) =x_{1}( u_{1}) +x_{2}( u_{2})
$ (where $\dim  x=\dim  x_{1}=\dim  x_{2}$), then $x( u) $ describes
a superposition (or mixture) of manifolds as shown in Figure \ref{XRef-Figure-822102129}b.
This type of manifold arises when the underlying degrees of freedom
are simultaneously measured by the same sensor. If there is little
or no overlap between $x_{1}( u_{1}) $ and $x_{2}( u_{2}) $ then
this is approximately equivalent to the case $x( u) =(x_{1}( u_{1})
,x_{2}( u_{2}) )$ (where $\dim  x=\dim  x_{1}+\dim  x_{2}$ and $\dim
u=\dim  u_{1}+\dim  u_{2}$). On the other hand, where there is a
significant amount of overlap so that $x_{1}( u_{1}) .x_{2}( u_{2})
>0$, there is no such correspondence. Assuming $\Pr ( u) =\Pr (
u_{1}) \Pr ( u_{2}) $ then $\Pr ( x) $ is given by $\Pr ( x) =\int
du_{1}du_{2} \Pr ( u_{1}) \Pr ( u_{2}) \delta ( x-x_{1}( u_{1})
-x_{2}( u_{2}) ) $.

More generally, $x( u) $ can be written as $x( u) =x( u_{1},u_{2})
$ where $x( u_{1},u_{2}) $ has no special dependence on $u_{1}$
and $u_{2}$. Although the manifolds are independently parameterised
by $u_{1}$ and $u_{2}$, when they are mapped to $x$ their tensor
product structure is disguised by the mapping function $x( u_{1},u_{2})
$ which is usually not invertible. The superposition of manifolds
$x( u) =x_{1}( u_{1}) +x_{2}( u_{2}) $ is a special case of this
effect.
\subsection{Mapping of Data Manifolds}\label{XRef-Section-822101823}

Given examples of the raw data $x$ how can an approximation to the
mapping function $x( u) $ be constructed? The detailed approach
will be described in Section \ref{XRef-Section-822101712},\ \ but
the basic geometric ideas will be described here. The basic idea
is to cut the manifold into pieces whilst retaining only a limited
amount of information about each piece, and then to reassemble these
pieces to reconstruct an approximation to the manifold. This process
is imperfect because it is disrupted by discarding some of the information
about each piece, so the reconstructed manifold is not a perfect
copy of the original manifold. This loss of information is critical
to the success of this process, because if perfect information were
retained then there would be no need to discover a clever way of
cutting the manifold into pieces, and thus no possibility of discovering
the structure of the manifold (e.g. whether it is a simple tensor
product). The information that is preserved depends on exactly how
the curved manifold is mapped to a new representation (see Section
\ref{XRef-Section-822101712} for details).
\begin{figure}[h]
\begin{center}
\includegraphics{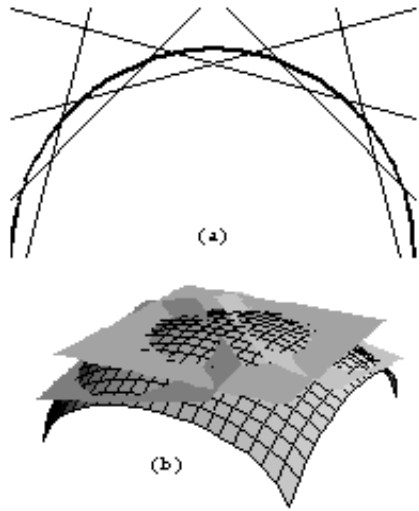}

\end{center}
\caption{Using hyperplanes to slice pieces off convex curved manifolds.
Slicing a curved manifold into pieces prepares it for mapping to
another representation. (a) 1-dimensional manifold with arcs being
sliced off by chords. (b) 2-dimensional manifold with caps being
sliced off by planes (only a few of these are shown in order to
keep the diagram simple).}\label{XRef-Figure-822102423}
\end{figure}

Figure \ref{XRef-Figure-822102423}a shows an example of how a convex
1-dimensional manifold can be cut into overlapping pieces by a set
of lines, and Figure \ref{XRef-Figure-822102423}b shows the generalisation
to the 2-dimensional case. This process is considerably simplified
if the manifold is convex because then the hyperplane slices off
a localised piece of the manifold, as required.
\begin{figure}[h]
\begin{center}
\includegraphics{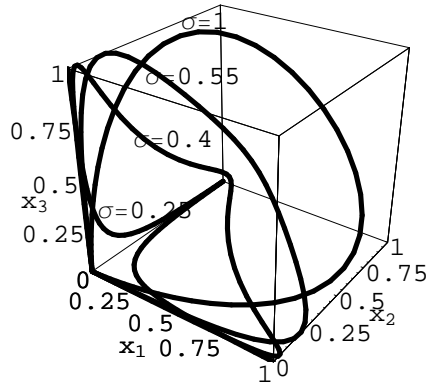}

\end{center}
\caption{Manifolds generated by a 1-dimensional object. The data
vector is $x=(\cdots ,x_{-2},x_{-1},x_{0},x_{1},x_{2},\cdots )$\ \ where
$x_{i}=\exp ( -\frac{(i-a)^{2}}{2\sigma ^{2}}) $, $\sigma $ is the
width of the object function, $a$ ($-\infty <a<\infty $) is the
position of the object, and $i$ ($i=0,\pm 1,\pm 2,\cdots $) is the
location of the points where the object amplitude is sampled. The
manifolds shown are 3-dimensional embeddings $(x_{1},x_{2},x_{3})$
of the 1-dimensional curved manifolds generated as $a$ varies for
a variety of object widths $\sigma $. For $\sigma =0.25$ (i.e. a
narrow object function) the manifold is concave with cusps, as $\sigma
$ is increased the concavity and the cusps become less pronounced
until the manifold crosses the border between being concave and
being convex, and for $\sigma =1$ (i.e. a wide object function)
the manifold is smoothly convex. Concave manifolds with cusps are
{\itshape not} well suited to being sliced apart by hyperplanes
whereas smoothly convex manifolds {\itshape are} well suited, and
this type of convex manifold is typical of high-dimensional data
which also has a high resolution so that each object covers several
sample points.}\label{XRef-Figure-822102611}
\end{figure}

Figure \ref{XRef-Figure-822102611} shows an example of how the convexity
assumption can break down. The full embedding space contains the
vector formed from an array of samples of a 1-dimensional object
function, but only three dimensions of the full embedding space
are shown in Figure \ref{XRef-Figure-822102611}. Several scenarios
are shown ranging from a narrow object (i.e. undersampled) to a
broad object (i.e. oversampled). Oversampling leads to a smooth
convex manifold, whereas undersampling leads to a concave manifold
with cusps. Typically, convex manifolds occur in signal and image
processing where the raw data are sampled at a high enough rate,
and non-convex manifolds occur when the raw data has already been
processed into a low-dimensional form, such as when some underlying
degrees of freedom (or features) have already been extracted from
the raw data.
\begin{figure}[h]
\begin{center}
\includegraphics{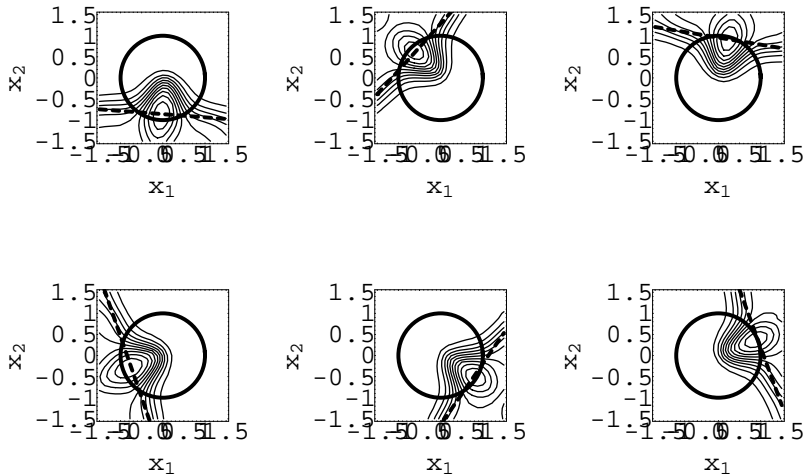}

\end{center}
\caption{Using a stochastic vector quantiser (SVQ) to map a curved
manifold to a new representation (see Section \ref{XRef-Section-822101712}
for details). The manifold is the unit circle which is softly sliced
up by the SVQ posterior probabilities, which are defined as the
(normalised) outputs of a set of sigmoid functions, which in turn
depend on a set of weight vectors and biases. For each sigmoid function
a dashed line is drawn to show where its (unnormalised) output is
$\frac{1}{2}$, although here it is the curved contours of the posterior
probabilities (rather than the dashed lines) that are actually used
to slice up the manifold.}\label{XRef-Figure-82210284}
\end{figure}

Figure \ref{XRef-Figure-82210284} shows the results obtained by
for a circular manifold using the stochastic vector quantiser (SVQ)
approach of Section \ref{XRef-Section-822101712}. The results correspond
to Figure \ref{XRef-Figure-822102423}a, except that now the slicing
is done softly in order to preserve additional information about
the manifold, and to ensure that the reconstructed manifold does
not show artefacts when the slices are reassembled.
\begin{figure}[h]
\begin{center}
\includegraphics{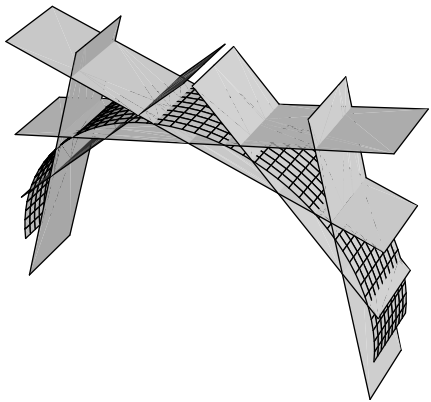}

\end{center}
\caption{Using hyperplanes to slice pieces off a curved manifold.
The manifold is 2-dimensional with a large and a small degree of
freedom. Each hyperplane slices through the manifold in such a way
that it cuts off a piece of the manifold that has a {\itshape limited}
range of values of the large degree of freedom but {\itshape all}
possible values of the small degree of freedom. This is the basic
means by which a manifold can be mapped to a new representation.}\label{XRef-Figure-822102942}
\end{figure}

Figure \ref{XRef-Figure-822102942} shows an example of how a convex
$(1+\epsilon )$-dimensional manifold (a small length extracted from
a cylindrical surface) can be cut into overlapping pieces by a set
of planes. The 1 in $(1+\epsilon )$ is a large degree of freedom
(arc length around the cylinder)), whereas the $\epsilon $ in $(1+\epsilon
)$ is a small degree of freedom (length along the cylinder) because
it has a small amplitude compared to the large degree of freedom.
Because of the orientation of the planes they are insensitive to
the small degree of freedom, so the reconstructed manifold is 1-dimensional
manifold (i.e. the $\epsilon $ component has been discarded). The
orientation of the planes may be used in various ways to control
their sensitivity to the manifold, and some quite sophisticated
examples of this will be discussed in Section \ref{XRef-Section-822101724}.
These results generalise to soft slicing as used in Figure \ref{XRef-Figure-82210284}.
\section{Learning a Manifold}\label{XRef-Section-822101712}

In order to learn how to represent the structure of a data manifold
a flexible framework needs to be used. In this paper an approach
will be used in which the manifold is mapped to a lower resolution
representation in such a way that a good approximation to the original
manifold can be reconstructed. Key requirements are that these mappings
can be cascaded to form sequences of representations of progressively
lower resolution having more and more invariance with respect to
details in the original manifold, and that the mappings can learn
to represent tensor products of manifolds so that the representation
of the manifold can split into separate channels. To achieve this
it is sufficient to use a variant \cite{Luttrell2003} of the standard
vector quantiser \cite{LindeBuzoGray1980} to gradually compress
the data.

In Section \ref{XRef-Section-822103446} the theory of stochastic
vector quantisers (SVQ) is presented, and in Section \ref{XRef-Section-822103459}
it is extended to chains of linked SVQs.
\subsection{Stochastic Vector Quantiser}\label{XRef-Section-822103446}

As was discussed in Section \ref{XRef-Section-82210171} a procedure
is needed for cutting a manifold into pieces and then reassembling
these pieces to reconstruct the manifold. It turns out that all
of the required properties emerge automatically from vector quantisers
(VQ), and their generalisation to stochastic vector quantisers (SVQ).
\begin{figure}[h]
\begin{center}
\includegraphics{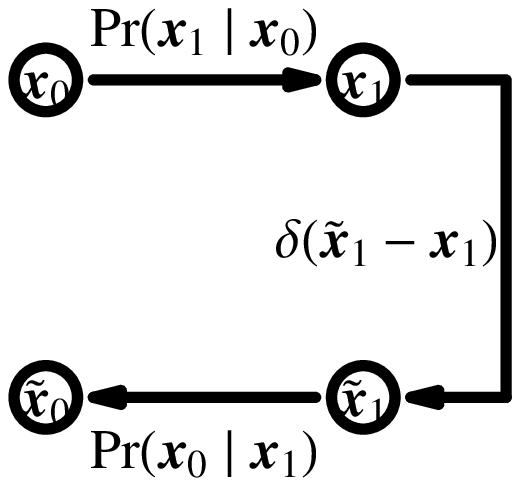}

\end{center}
\caption{A matched encoder/decoder pair represented as a folded
Markov chain (FMC) $x_{0}\longrightarrow x_{1}\longrightarrow \overset{~}{x}_{1}\longrightarrow
\overset{~}{x}_{0}$. The input $x_{0}$ is encoded as $x_{1}$ which
is then passed along a distortionless communication channel to become
$\overset{~}{x}_{1}$ which is then decoded as $\overset{~}{x}_{0}$.
The encoder is modelled using the conditional probability $\Pr (
x_{1}|x_{0}) $ to allow for the possibility that the encoder is
stochastic, and the corresponding decoder is modelled using the
Bayes inverse conditional probability $\Pr ( \overset{~}{x}_{0}|\overset{~}{x}_{1})
$. The distortionless communication channel is modelled using the
delta function $\delta ( \overset{~}{x}_{1}-x_{1}) $.}\label{XRef-Figure-822103925}
\end{figure}

Figure \ref{XRef-Figure-822103925} shows a folded Markov chain (FMC)
as described in \cite{Luttrell1994}. An FMC encodes its input $x_{0}$
(i.e. cuts the input manifold into pieces using the conditional
PDF $\Pr ( x_{1}|x_{0}) $) and then reconstructs an approximation
to its input $\overset{~}{x}_{0}$ (i.e. reassembling the pieces
to reconstruct the input manifold using the Bayes inverse PDF $\Pr
( \overset{~}{x}_{0}|x_{1}) =\frac{\Pr ( x_{1}|\overset{~}{x}_{0})
\Pr ( \overset{~}{x}_{0}) }{\Pr ( x_{1}) }$), so it is ideally suited
to the task at hand. An objective function $D$ needs to be defined
to measure how accurately the reconstruction $\overset{~}{x}_{0}$
approximates the original input $x_{0}$.

It is simplest to use a Euclidean objective function that measures
the average squared (i.e. $L^{2}$) distance $\|x_{0}-\overset{~}{x}_{0}\|^{2}$,
and which must be minimised with respect to the encoder $\Pr ( x_{1}|x_{0})
$ (note that the decoder $\Pr ( \overset{~}{x}_{0}|\overset{~}{x}_{1})
$ is then completely determined by Bayes' theorem).
\begin{equation}
D=\int dx_{0}dx_{1}d\overset{~}{x}_{0}d\overset{~}{x}_{1}\Pr ( x_{0})
\Pr ( x_{1}|x_{0}) \delta ( \overset{~}{x}_{1}-x_{1}) \Pr ( \overset{~}{x}_{0}|\overset{~}{x}_{1})
\left\| x_{0}-\overset{~}{x}_{0}\right\| ^{2}%
\label{XRef-Equation-822104133}
\end{equation}

\noindent Using Bayes' theorem Equation \ref{XRef-Equation-822104133}
can be manipulated into the form \cite{Luttrell1994}
\begin{equation}
D=2\int dx_{0}dx_{1}\Pr ( x_{0}) \Pr ( x_{1}|x_{0}) \left\| x_{0}-\overset{~}{x}_{0}(
x_{1}) \right\| ^{2}%
\label{XRef-Equation-822104220}
\end{equation}

\noindent where $D$ must be minimised with respect to both the encoder
$\Pr ( x_{1}|x_{0}) $ and the reconstruction vector $\overset{~}{x}_{0}(
x_{1}) $. Note that this simplification of Equation \ref{XRef-Equation-822104133}
into Equation \ref{XRef-Equation-822104220} depends critically on
the Euclidean form of the objective function.
\begin{figure}[h]
\begin{center}
\includegraphics{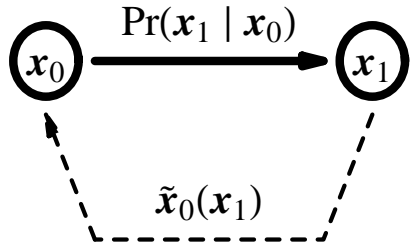}

\end{center}
\caption{An encoder/decoder pair represented as the chain $x_{0}\longrightarrow
x_{1}\longrightarrow \overset{~}{x}_{0}( x_{1}) $. This contains
only those parts of the FMC that affect the Euclidean distortion
objective function.}\label{XRef-Figure-82210444}
\end{figure}

Figure \ref{XRef-Figure-82210444} is a transformed version of Figure
\ref{XRef-Figure-822103925} that reflects the transformation of
Equation \ref{XRef-Equation-822104133} into Equation \ref{XRef-Equation-822104220}.
The encoder $\Pr ( x_{1}|x_{0}) $ is the most important part of
this diagram, whereas the reconstruction vector $\overset{~}{x}_{0}(
x_{1}) $ is less important so it is shown as a dashed line.

Thus far a non-parametric representation of $\Pr ( x_{1}|x_{0})
$ and $\overset{~}{x}_{0}( x_{1}) $ has been used, so analytic minimisation
of $D$ \cite{Luttrell1994} leads to $\Pr ( x_{1}|x_{0}) \longrightarrow
\delta ( x_{1}-x_{1}( x_{0}) ) $, in which case the encoder could
be perfect (i.e. lossless) and it would not be possible to discover
the structure of the input manifold, as discussed in Section \ref{XRef-Section-82210171}.
To make progress constrained forms of $\Pr ( x_{1}|x_{0}) $ and
$\overset{~}{x}_{0}( x_{1}) $ must be used in order to limit the
resources available to the encoder/decoder, and thus force it to
discover clever ways of mapping the input manifold to reduce the
damage caused by having only limited coding resources.

One way of constraining the encoder/decoder is for $x_{1}$ to be
a {\itshape scalar} index $y_{1}$ where $y_{1}=1,2,\cdots ,m_{1}$
($m_{1}$ is the size of the code book), which is a {\itshape single}
sample drawn from the encoder $\Pr ( x_{1}|x_{0}) $. Analytic minimisation
of $D$ \cite{Luttrell1994} now leads to $\Pr ( x_{1}|x_{0}) \longrightarrow
\delta _{y_{1},y_{1}( x_{0}) }$ so that $D=2\int dx_{0}\Pr ( x_{0})
\|x_{0}-\overset{~}{x}_{0}( y_{1}( x_{0}) ) \|^{2}$ which is the
objective function for a standard least squares vector quantiser
\cite{LindeBuzoGray1980}.

A better way of constraining the encoder/decoder is for $x_{1}$
to be the {\itshape histogram} $(\nu _{1},\nu _{2},\cdots ,\nu _{m_{1}})$
of counts of independent samples of the scalar index $y_{1}$ ($n_{1}=\sum
_{y_{1}=1}^{m_{1}}\nu _{y_{1}}$ is the total number of samples),
and for $\overset{~}{x}_{0}( x_{1}) $ to be approximated as $\overset{~}{x}_{0}(
x_{1}) \approx \sum _{y_{1}=1}^{m_{1}}\frac{\nu _{y_{1}}}{n_{1}}\overset{~}{x}_{0}(
y_{1}) $ (rather than using the full functional form $\overset{~}{x}_{0}(
\nu _{1},\nu _{2},\cdots ,\nu _{m_{1}}) $). Although it is still
possible to obtain analytic results it usually requires a lot of
calculation \cite{Luttrell1999a}, and it is generally better to
use a numerical optimisation approach. An upper bound for the objective
function $D$ is then given by \cite{Luttrell1997}
\begin{equation}
\begin{array}{rl}
 D & \leq \frac{2}{n_{1}}\int dx_{0}\Pr ( x_{0}) \sum \limits_{y_{1}=1}^{m_{1}}\Pr
( y_{1}|x_{0}) \left\| x_{0}-\overset{~}{x}_{0}( y_{1}) \right\|
^{2} \\
  &  \begin{array}{cc}
   &
\end{array}+\frac{2\left( n_{1}-1\right) }{n_{1}}\int dx_{0}\Pr
( x_{0}) \left\| x_{0}-\sum \limits_{y_{1}=1}^{m_{1}}\Pr ( y_{1}|x_{0})
\overset{~}{x}_{0}( y_{1}) \right\| ^{2}
\end{array}%
\label{XRef-Equation-822104850}
\end{equation}

\noindent where the unconstrained encoder/decoder corresponds to
the left hand side of Equation \ref{XRef-Equation-822104850} and
the constrained encoder/decoder corresponds to the right hand side
of Equation \ref{XRef-Equation-822104850}. Note how the random fluctuations
in the {\itshape multiple} sample histogram are analytically summed
over in Equation \ref{XRef-Equation-822104850}, leaving only the
{\itshape single} sample encoder $\Pr ( y_{1}|x_{0}) $ to be optimised.

A further constraint is to assume that $\Pr ( y_{1}|x_{0}) $ is
parameterised as the normalised output of a set of sigmoid functions
\begin{equation}
\begin{array}{rl}
 \Pr ( y_{1}|x_{0})  & =\frac{Q( y_{1}|x_{0}) }{\sum _{y_{1}^{\prime
}=1}^{m_{1}}Q( y_{1}^{\prime }|x_{0}) } \\
 Q( y_{1}|x_{0})  & =\frac{1}{1+\exp ( -w_{10}( y_{1}) .x_{0}-b_{1}(
y_{1}) ) }
\end{array}%
\label{XRef-Equation-82210519}
\end{equation}

\noindent where $Q( y_{1}|x_{0}) $ is the unnormalised output from
code index $y_{1}$, depending on the weight vector $w_{10}( y_{1})
$ and the bias $b_{1}( y_{1}) $. This parameterisation of $\Pr (
y_{1}|x_{0}) $ ensures that it can be used to slice pieces off convex
manifolds as illustrated in Figure \ref{XRef-Figure-82210284}. Optimisation
of the objective function is then achieved by gradient descent variation
of the three sets of parameters $w_{10}( y_{1}) $, $b_{1}( y_{1})
$, and $\overset{~}{x}_{0}( y_{1}) $. These (and other) derivatives
of the objective function were given in \cite{Luttrell1997}.

The constrained objective function in Equation \ref{XRef-Equation-822104850}
and Equation \ref{XRef-Equation-82210519} yields a great variety
of useful results, such as the simple result shown in Figure \ref{XRef-Figure-82210284}
which used $m_{1}=6$, $n_{1}=20$, and $x_{0}=(\cos  \theta ,\sin
\theta )$ with $\theta $ uniformly distributed in $[0,2\pi ]$, to
learn a mapping from the 1-dimensional input manifold embeded in
a 2-dimensional space ($\dim  x_{0}=2$) to a 6-dimensional space
($m_{1}=6$). This is the key objective function that can be used
to optimise the mapping of the input manifold to a new representation
$\Pr ( y_{1}|x_{0}) $ for $y_{1}=1,2,\cdots ,m_{1}$. Minimising
the Euclidean distortion ensures that $\Pr ( y_{1}|x_{0}) $ defines
an optimal mapping of the input manifold, such that when $n_{1}$
samples are drawn from $\Pr ( y_{1}|x_{0}) $ they contain enough
information to form an accurate reconstruction of $x_{0}$.
\begin{figure}[h]
\begin{center}
\includegraphics{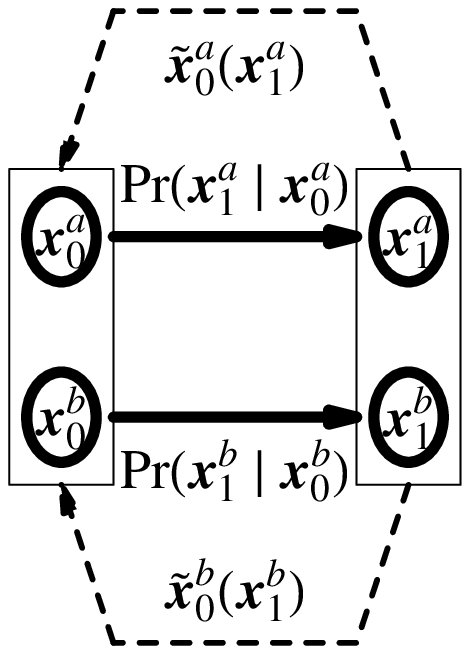}

\end{center}
\caption{A factorial encoder/decoder pair represented as the pair
of disconnected chains $x_{0}^{a}\longrightarrow x_{1}^{a}\longrightarrow
\overset{~}{x}_{0}^{a}( x_{1}^{a}) $ and $x_{0}^{b}\longrightarrow
x_{1}^{b}\longrightarrow \overset{~}{x}_{0}^{b}( x_{1}^{b}) $. The
input vector is $x_{0}=(x_{0}^{a},x_{0}^{b})$ highlighted by the
left hand rectangle, the code is $x_{1}=(x_{1}^{a},x_{1}^{b})$ highlighted
by the right hand rectangle, and the reconstruction is $\overset{~}{x}_{0}=(\overset{~}{x}_{0}^{a},\overset{~}{x}_{0}^{b})$.
The dependencies amongst the variables is indicated by the arrows
in the diagram, which shows that subspaces $a$ and $b$ are {\itshape
independently} encoded/decoded.}\label{XRef-Figure-822105244}
\end{figure}

Figure \ref{XRef-Figure-822105244} is a transformed version of Figure
\ref{XRef-Figure-82210444} that shows an example of the structure
of some types of optimal solution that are obtained by minimising
the constrained objective function in Equation \ref{XRef-Equation-822104850}.
The tensor product structure of the input manifold is revealed in
this type of solution, because the input vector $x_{0}$ splits into
two parts as $x_{0}=(x_{0}^{a},x_{0}^{b})$ each of which is separately
encoded/decoded. This type of factorial encoder is favoured by limiting
the size of the code book $m$ and by using an intermediate number
of samples $n_{1}$ ($n_{1}=1$ leads to a standard VQ, and $n_{1}\longrightarrow
\infty $ allows too many coding resources to lead to clever encoding
schemes).

The self-organised emergence of factorial encoders is one of the
major strengths of the SVQ approach. It allows the code book to
split into two or more separate smaller code books in a data driven
way rather than being hard-wired into the code book at the outset
(e.g. \cite{RossZemel2003}).
\subsection{Chain of Stochastic Vector Quantisers}\label{XRef-Section-822103459}

The encoder/decoder in Figure \ref{XRef-Figure-82210444} leads to
useful results for mapping the input data manifold when its operation
is constrained in various ways. A much larger variety of mappings
may be constructed if the encoder/decoder is viewed as a basic module,
and then networks of linked modules are used to process the data
\cite{Luttrell2003}. It is simplest to regard this type of network
as progressively mapping the input manifold as it flows through
the network modules.
\begin{figure}[h]
\begin{center}
\includegraphics{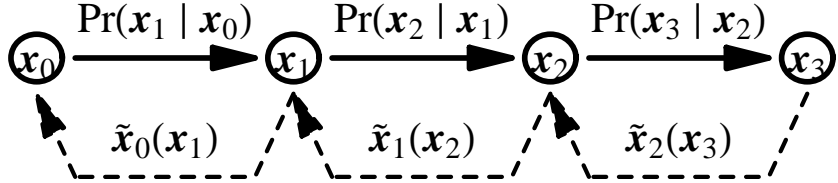}

\end{center}
\caption{A 3-stage chain of linked SVQs. The $l^{\textup{th}}$ encoder
is modelled using the conditional probability $\Pr _{l,l-1}( x_{l}|x_{l-1})
$, and the corresponding decoder is modelled using the reconstruction
vectors $\overset{~}{x}_{l-1}( x_{l}) $, where each $x_{l}$ is a
histogram of samples.}\label{XRef-Figure-822105530}
\end{figure}

Figure \ref{XRef-Figure-822105530} shows a 3-stage chain of linked
encoder/decoders of the type shown in Figure \ref{XRef-Figure-82210444}.
The important part of this diagram is the processing chain which
is the solid line flowing from left to right at the top of the diagram
creating the Markov chain $x_{0}\overset{\Pr ( x_{1}|x_{0}) }{\longrightarrow
}x_{1}\overset{\Pr ( x_{2}|x_{1}) }{\longrightarrow }x_{2}\overset{\Pr
( x_{3}|x_{1}) }{\longrightarrow }x_{3}$. The reconstruction vectors
$\overset{~}{x}_{l-1}( x_{l}) $ for $l=1,2,3$ are the dashed lines
flowing from right to left.

The state $x_{l}$ of layer $l$ of the chain is the histogram $(\nu
_{1},\nu _{2},\cdots ,\nu _{m_{l}})$ of counts of samples drawn
from $\Pr ( y_{l}|x_{l-1}) $ ($n_{l}=\sum _{y_{l}=1}^{m_{l}}\nu
_{y_{l}}$ is the total number of samples). In numerical implementations
$x_{l}$ is chosen to be the (normalised) histogram for an infinite
number of samples (i.e. the relative frequencies implied by $\Pr
( y_{l}|x_{l-1}) $), and $\overset{~}{x}_{l-1}( x_{l}) $ is chosen
to depend on only a finite number $m_{l}$ of samples randomly selected
from this histogram\ \ $\overset{~}{x}_{l-1}( x_{l}) \approx \sum
_{y_{l}=1}^{m_{l}}\frac{\nu _{y_{l}}}{n_{l}}\overset{~}{x}_{l-1}(
y_{l}) $. This choice of how to operate the network is not unique
but it has the advantage of\ \ simplifying the computations. The
{\itshape infinite} number of samples used in $x_{l}$ ensures that
the $x_{l}$ do not randomly fluctuate, so no Monte Carlo simulations
are required to implement the feed-forward flow through the network.
The {\itshape finite} number of samples $n_{l} $ used in $\overset{~}{x}_{l-1}(
x_{l}) $ leads to exactly the same objective function as in Equation
\ref{XRef-Equation-822104850} where the random fluctuations are
analytically summed over, which ensures that each decoder has limited
resources and thus forces the optimisation of the network to discover
intelligent ways of encoding the data. This type of network reduces
to a standard way of using a Markov chain when only a single sample
is drawn from each of the $\Pr ( y_{l}|x_{l-1}) $.

Each stage of the chain corresponds to an objective function of
the form shown in Equation \ref{XRef-Equation-822104850} but applied
to the $l^{\textup{th}}$ stage of the chain. The total objective
function is a weighted sum of these individual contributions. This
encourages all of the mappings in the chain to minimise their average
Euclidean reconstruction error, which gives a progressive mapping
of the input manifold along the processing chain. However, the relative
weighting of the later stages of the chain must not be too great
otherwise they force the earlier mappings in the chain to become
singular (e.g. all inputs mapped to the same output), because the
output of a singular mapping can be mapped with little or no more
contribution to the overall objective function further along the
chain. A less extreme form of this phenomenon can be used to encourage
factorial encoders to emerge, because they produce a (normalised)
histogram output state that has a smaller volume (in the Euclidean
sense) than a non-factorial encoder, which reduces the size of the
contribution to the overall objective function from the next stage
in the chain.

If the chain network topology in Figure \ref{XRef-Figure-822105530}
is combined with the factorial encoder/decoder property of SVQs
shown in Figure \ref{XRef-Figure-822105244} then all acyclic network
topologies are possible. This can be seen intuitively because flow
through the chain corresponds to flow along the time-like direction
in an acyclic network (i.e. following the directed links), and multiple
parallel branches occur wherever there is an SVQ factorial encoder
in the chain. An example of the emergence of this type of network
topology will be shown in Section \ref{XRef-Section-822101724}.
\section{Learning a Hierarchical Network}\label{XRef-Section-822101724}

The purpose of this section is to demonstrate the self-organised
emergence of a hierarchical network topology starting from a chain-like
topology of the type shown in Figure \ref{XRef-Figure-822105530}.
For the purpose of this demonstration the raw data must have an
appropriate correlation structure, which will be achieved by generating
the data as a set of hierarchically correlated phases. Thus each
data vector is a 4-dimensional vector of phases $\phi =(\phi _{1},\phi
_{2},\phi _{2},\phi _{4})$, where the $\phi _{i}$ are the leaf nodes
of a binary tree of phases, where the binary splitting rule used
is $\phi \longrightarrow (\phi -\alpha ,\phi +\beta )$ with $\alpha$
and $\beta$ being independently and uniformly sampled from the interval
$[0,\frac{\pi }{2}]$, and the phase of the root node is uniformly
distributed in the interval $[0,2\pi ]$. This will lead to each
of the $\phi _{i}$ being uniformly distributed phase variables thus
uniformly occupying a circular manifold. However, because the $\phi
_{i}$ are correlated due to the way that they are generated by the
binary splitting process, the $\phi $ do {\itshape not} uniformly
populate a 4-torus manifold (i.e. tensor product of 4 circles).

Figure \ref{XRef-Figure-822102942} showed an example of what the
manifold of a pair of correlated variables looks like, with a large
degree of freedom (e.g. $\phi _{1}+\phi _{2}$) and a small degree
of freedom (e.g. $\phi _{1}-\phi _{2}$), and the hyperplanes encoding
the manifold in such a way as to discard information about the small
degree of freedom (e.g. $\phi _{1}-\phi _{2}$). In this way the
3-stage chain in Figure \ref{XRef-Figure-822105530} can progressively
discard information about small degrees of freedom in $\phi $, starting
with a 4-torus manifold (non-uniformly populated) and ending up
with a circular manifold (uniformly populated), as will be seen
below. This is the basic idea behind using this type of self-organising
network for data fusion.
\begin{figure}[h]
\begin{center}
\includegraphics{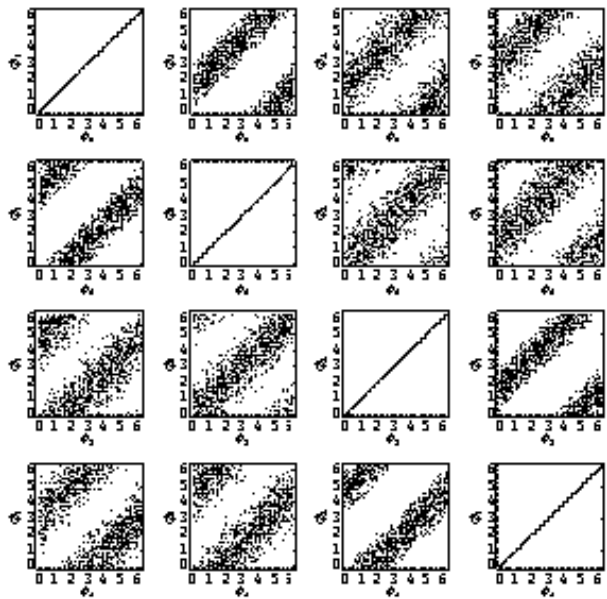}

\end{center}
\caption{ Co-occurrence matrices of all pairs of phases. Note that
the block structure is symmetric so each off-diagonal co-occurrence
matrix appears twice. The hierarchical correlations cause $\phi
_{1}$and $\phi _{2}$\ \ (and similarly $\phi _{3}$ and $\phi _{4}$)
to be more strongly correlated with each other than $\phi _{2}$
and $\phi _{3}$.}\label{XRef-Figure-822105946}
\end{figure}

Figure \ref{XRef-Figure-822105946} shows the co-occurrence matrices
of pairs of the $\phi _{i}$ displayed as scatter plots. The bands
in these plots wrap around circularly and correspond to the manifold
shown in Figure \ref{XRef-Figure-822102942}. Because $\phi _{1}$
and $\phi _{2}$ (and also $\phi _{3}$ and $\phi _{4}$) lie close
to each other in the hierarchy, $\Pr ( \phi _{1,}\phi _{2}) $ and
$\Pr ( \phi _{3},\phi _{4}) $ have a narrower band than $\Pr ( \phi
_{1,}\phi _{3}) $, $\Pr ( \phi _{1,}\phi _{4}) $, $\Pr ( \phi _{2,}\phi
_{3}) $ and $\Pr ( \phi _{2,}\phi _{4}) $.

A 3-stage chain of linked SVQs of the type shown in Figure \ref{XRef-Figure-822105530}
is now trained, where each stage contributes an objective function
of the form shown in Equation \ref{XRef-Equation-822104850} and
Equation \ref{XRef-Equation-82210519}. The sizes $M$ of each of
the 4 network layers are $M=(8,16,8,4)$. The size of layer 0 (the
input layer) is determined by the dimensionality of the input data,
whereas the sizes of each of the other layers is chosen to be 4
times the number of phase variables that each is expected to use
in its encoding of the input data, which encourages the progressive
removal of small degrees of freedom from the data as it flows along
the chain into ever smaller layers. The number of samples $n$ used
for each of the 3 SVQ stages are $n=(20,20,20)$, which are large
enough to allow each SVQ to develop into a factorial encoder, so
that the processing can proceed in parallel along several paths
(which are progressively fused) along the chain. The relative weightings
$\lambda $ assigned to the objective functions contributed by each
of the 3 SVQ stages are $\lambda =(1,5,0.1)$, where a large weighting
is assigned to the stage 2 SVQ to encourage the stage 1 SVQ to develop
into a factorial encoder, and a small weighting is assigned to the
stage 3 SVQ because the stage 2 SVQ needs no additional encouragement
to develop into a factorial encoder.

The network was trained by a gradient descent on the overall network
objective function, using a step size chosen separately for each
SVQ stage and separately for each of the 3 parameter types $w_{l,l-1}(
y_{l}) $, $b_{l}( y_{l}) $, and $\overset{~}{x}_{l-1,l}( y_{l})
$ in each SVQ stage (for $l=1,2,3$). The size of all of these step
size parameters was chosen to be large at the start of the training
schedule, and then gradually reduced as training progressed, with
the relative rate of reduction being chosen to encourage earlier
SVQ stages to converge before later SVQ stages. In general, different
choices of network parameters and training conditions lead to different
types of trained network, and since there is no prior reason for
choosing one particular solution in preference to another the choice
must be left up to the user. All the components of the weight vectors,
biases, and reconstruction vectors are initialised to random numbers
uniformly distributed in the interval\ \ $[-0.1,0.1]$.
\begin{figure}[h]
\begin{center}
\includegraphics{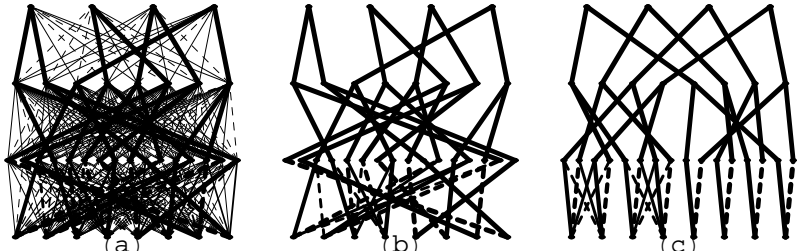}

\end{center}
\caption{ Reconstruction vectors $\overset{~}{x}_{l-1,l}( y_{l})
$ (for $l=1,2,3$) after training a 3-stage network of linked SVQs
on the hierarchically correlated phases data. These diagrams are
rotated $90^{\circ }$ anticlockwise relative to Figure \ref{XRef-Figure-822105530},
so the processing chain runs from bottom to top of each diagram.
Line thickness indicates the size of a reconstruction vector component,
and dashing indicates that the component is negative. (a) All reconstruction
vector components. (b) Largest reconstruction vector components
obtained by applying a threshold to the magnitude of each component.
(c) The same as (b) except that the positions of the nodes in all
layers (other than the input layer) have been permuted (along with
the connections between layers) to make the network topology clearer.}\label{XRef-Figure-82211324}
\end{figure}

\begin{figure}[h]
\begin{center}
\includegraphics{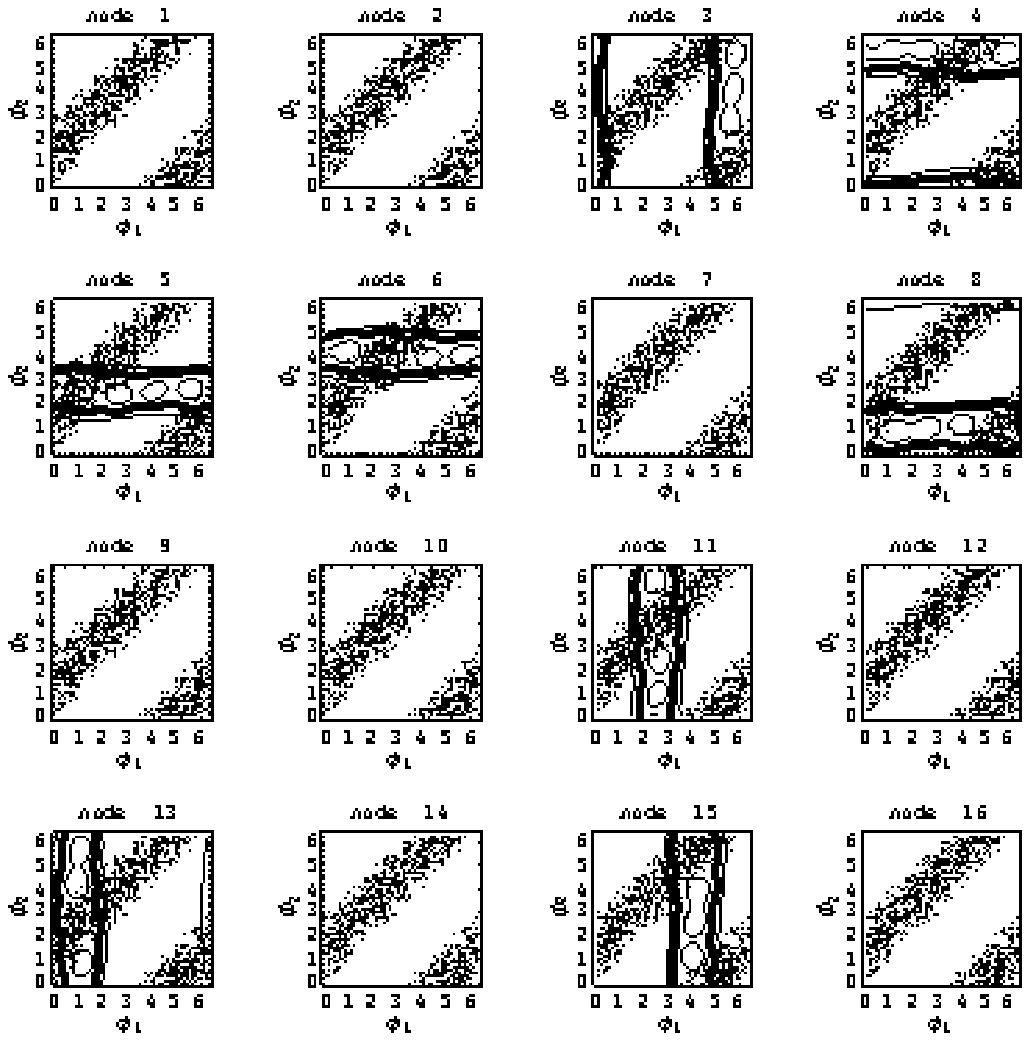}

\end{center}
\caption{ Node activities in layer 1 as a function of the inputs
$\phi _{1}$ and $\phi _{2}$ (with $\phi _{3}=\phi _{4}=0$) which
has the properties of a factorial encoder. In each plot the contours
representing the node reponse are overlaid on the co-occurrence
matrix of the pair of inputs $\phi _{1}$ and $\phi _{2}$. One of
the contour heights is drawn bold to highlight the region where
the node response is large. Half of the nodes do not respond at
all, and the other half split into two subsets of equal size, one
with high resolution in $\phi _{1}$ but completely insensitive to
$\phi _{2}$, and the other with high resolution in $\phi _{2}$ but
completely insensitive to $\phi _{1}$.\ \ The response is sensitive
to the small degree of freedom measured {\itshape across} the band
of the co-occurrence matrix. The non-zero responses correspond to
the 8 nodes in layer 1 that are strongly connected to $\phi _{1}$
and $\phi _{2}$.}\label{XRef-Figure-822111559}
\end{figure}

\begin{figure}[h]
\begin{center}
\includegraphics{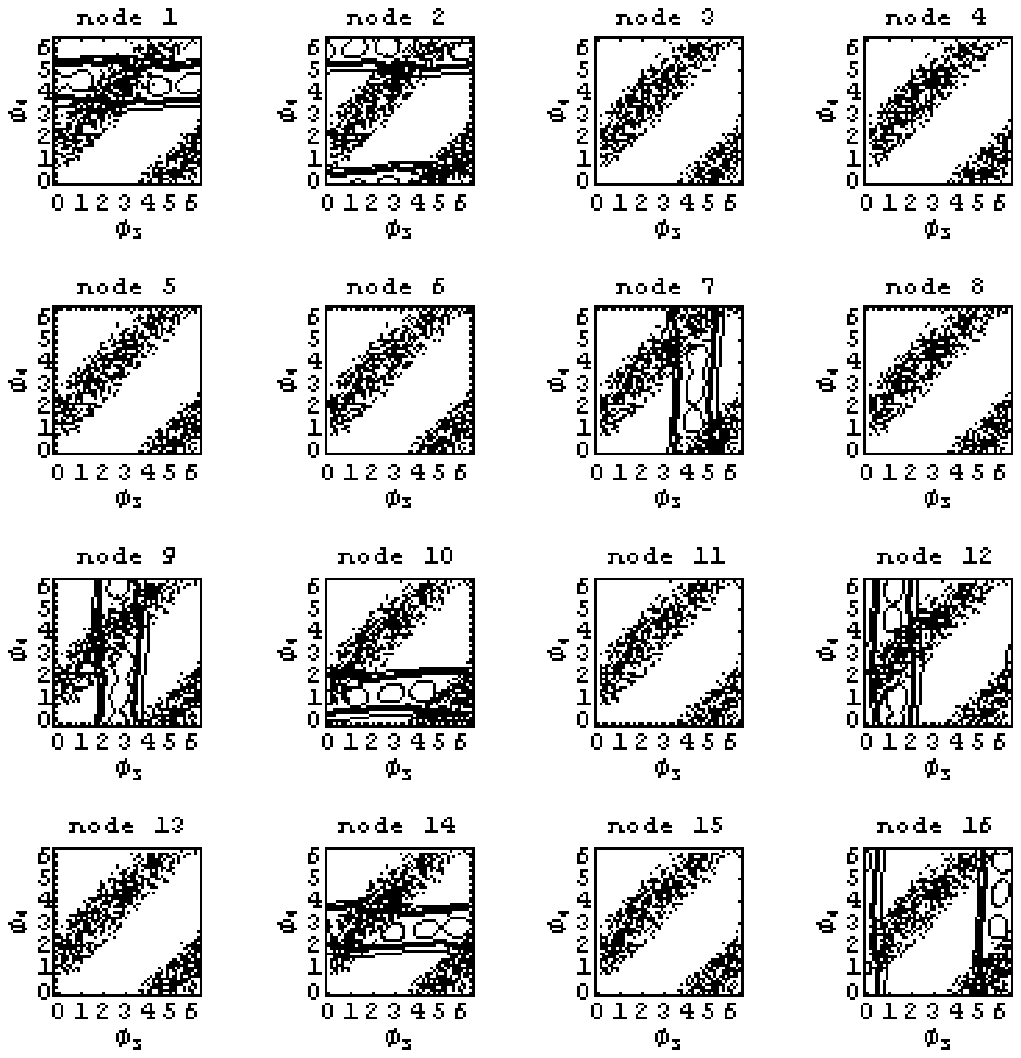}

\end{center}
\caption{ Node activities in layer 1 as a function of the inputs
$\phi _{3}$ and $\phi _{4}$ (with $\phi _{1}=\phi _{2}=0$) which
has the properties of a factorial encoder. The non-zero responses
correspond to the 8 nodes in layer 1 that are strongly connected
to $\phi _{3}$ and $\phi _{4}$.}\label{XRef-Figure-82211165}
\end{figure}

\begin{figure}[h]
\begin{center}
\includegraphics{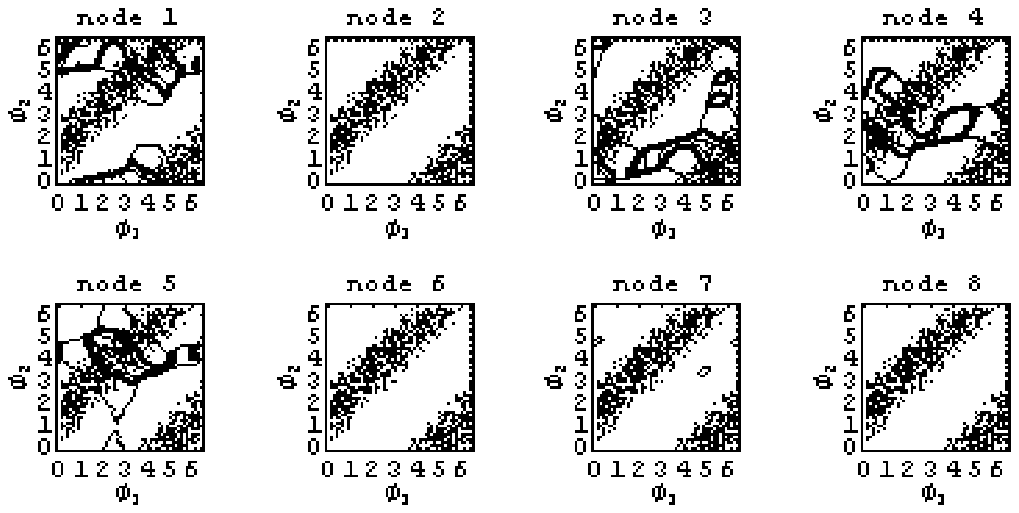}

\end{center}
\caption{Node activities in layer 2 as a function of the inputs
$\phi _{1}$ and $\phi _{2}$ (with $\phi _{3}=\phi _{4}=0$) which
has the properties of an invariant encoder. Half of the nodes do
not respond at all, and the other half respond to well-defined regions
in $\phi _{1}$ and $\phi _{2}$. The contours representing the node
response are overlaid on the co-occurrence matrix of the pair of
inputs $\phi _{1}$ and $\phi _{2}$. One of the contour heights is
drawn bold to highlight the region where the node response is large.
This shows that each node responds to a local region of the {\itshape
populated} region of the co-occurrence matrix, and to a limited
extent generalises outside this region. The response is invariant
with respect to the small degree of freedom measured {\itshape across}
the band of the co-occurrence matrix, which demonstrates that layer
2 has acquired an invariance that was absent in layer 1. There are
also non-zero responses in the {\itshape unpopulated} region of
the co-occurrence matrix which arise because the sum of the node
activities is normalised. The non-zero responses correspond to the
4 nodes in layer 2 that are strongly connected to $\phi _{1}$ and
$\phi _{2}$.}\label{XRef-Figure-822111928}
\end{figure}

\begin{figure}[h]
\begin{center}
\includegraphics{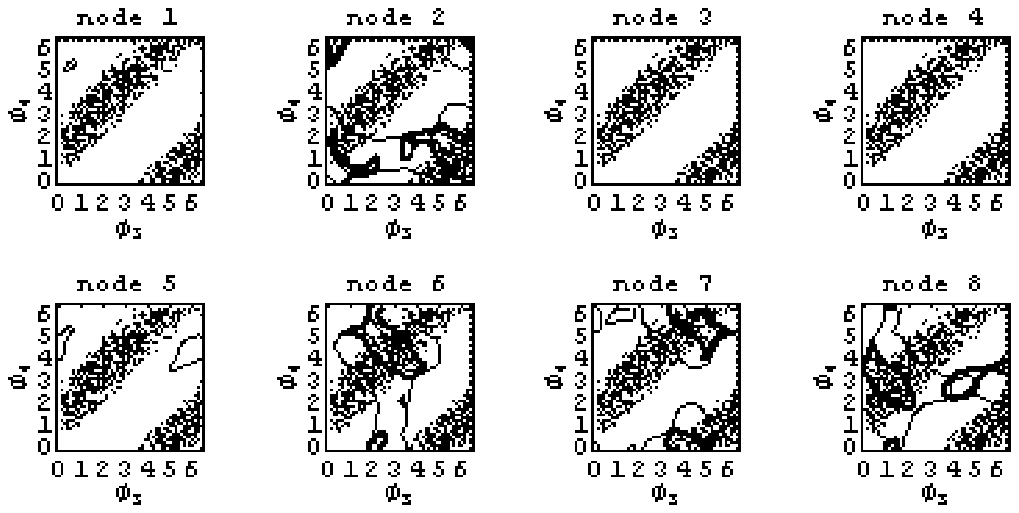}

\end{center}
\caption{ Node activities in layer 2 as a function of the inputs
$\phi _{3}$ and $\phi _{4}$ (with $\phi _{1}=\phi _{2}=0$) which
has the properties of an invariant encoder. The non-zero responses
correspond to the 4 nodes in layer 2 that are strongly connected
to $\phi _{3}$ and $\phi _{4}$.}\label{XRef-Figure-822111935}
\end{figure}

\begin{figure}[h]
\begin{center}
\includegraphics{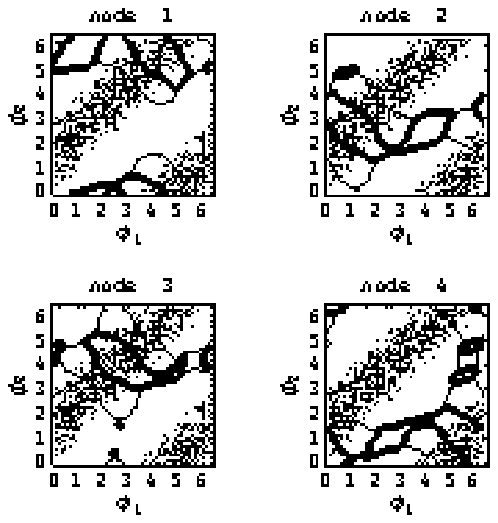}

\end{center}
\caption{ Node activities in layer 3 as a function of the inputs
$\phi _{1}$ and $\phi _{2}$ (with $\phi _{3}=\phi _{4}=0$) which
has the properties of an invariant encoder.}\label{XRef-Figure-82211249}
\end{figure}

\begin{figure}[h]
\begin{center}
\includegraphics{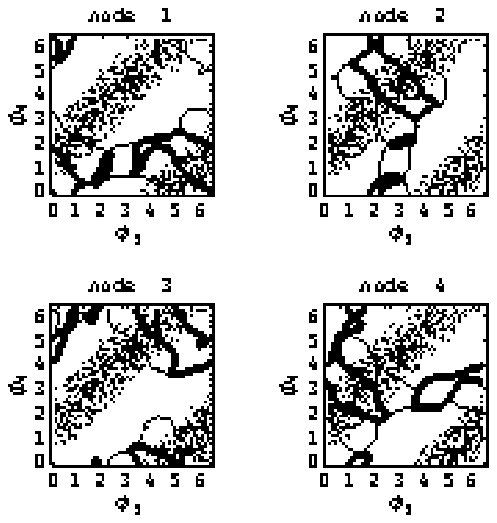}

\end{center}
\caption{ Node activities in layer 3 as a function of the inputs
$\phi _{3}$ and $\phi _{4}$ (with $\phi _{1}=\phi _{2}=0$) which
has the properties of an invariant encoder.}\label{XRef-Figure-822112415}
\end{figure}

Figure \ref{XRef-Figure-82211324} shows the reconstruction vectors
$\overset{~}{x}_{l-1,l}( y_{l}) $ (for $l=1,2,3$) in a trained 3-stage
chain of linked SVQs of the type shown in Figure \ref{XRef-Figure-822105530}.
Reconstruction vectors are displayed because they are easier to
interpret than weight vectors. The data $\phi =(\phi _{1},\phi _{2},\phi
_{2},\phi _{4})$ is embedded in an 8-dimensional input space as
$x=(x_{1},x_{2},x_{3},x_{4},x_{5},x_{6},x_{7},x_{8})$ where $(x_{2i-1},x_{2i})=(\cos
\phi _{i},\sin  \phi _{i})$. The key diagram is Figure \ref{XRef-Figure-82211324}c
which shows the largest components of the reconstruction vectors,
and has been reordered to make the hierarchical network topology
clear. Each of the first two stages of this network has learnt to
operate as two or more encoder/decoders (i.e. a factorial encoder/decoder)
as in Figure \ref{XRef-Figure-822105244}. The first stage of the
network breaks into 4 encoder/decoders that encode each of the $\phi
_{i}$ (see the results in Figure \ref{XRef-Figure-822111559} and
Figure \ref{XRef-Figure-82211165} for justification of this), the
second stage of the network breaks into 2 encoder/decoders that
encode $\phi _{1}+\phi _{2}$ and $\phi _{3}+\phi _{4}$ (see the
results in Figure \ref{XRef-Figure-822111928} and Figure \ref{XRef-Figure-822111935}
for justification of this), and the third stage of the network is
a single encoder that encodes $\phi _{1}+\phi _{2}+\phi _{3}+\phi
_{4}$ (see the results in Figure \ref{XRef-Figure-82211249} and
Figure \ref{XRef-Figure-822112415} for justification of this). The
connectivity in the stage 1 SVQ is not the same for all of the $\phi
_{i}$ because of the interaction between the thresholding prescription
used to create Figure \ref{XRef-Figure-82211324} and the different
orientation of each of the 4 parts of the stage 1 factorial encoder
with respect to each of the 4 corresponding circular input manifolds.

Although the network in Figure \ref{XRef-Figure-82211324} computes
using continuous-valued numbers, the thresholded reconstruction
vectors in Figure \ref{XRef-Figure-82211324}b may be inspected to
reveal the symbolic logic expressions that approximate to each of
the (thresholded) outputs $O_{i}( x) $ for $i=1,2,3,4$ from the
highest layer of the network (using logical negation $\overset{\_}{x}_{i}$
to denote $-x_{i}$ because the inputs lie in the range $[-1,1]$).
\begin{equation}
\begin{array}{rl}
 O_{1}( x)  & =\overset{\_}{x}_{2}\cap x_{3}\cap x_{5}\cap x_{8}
\\
 O_{2}( x)  & =x_{2}\cap \overset{\_}{x}_{3}\cap \overset{\_}{x}_{5}\cap
\overset{\_}{x}_{8} \\
  & =\overset{\_}{O}_{1}( x)  \\
 O_{3}( x)  & =\overset{\_}{x}_{1}\cap \overset{\_}{x}_{4}\cap \overset{\_}{x}_{6}\cap
x_{7} \\
 O_{4}( x)  & =x_{1}\cap x_{4}\cap x_{6}\cap \overset{\_}{x}_{7}
\\
  & =\overset{\_}{O}_{3}( x)
\end{array}
\end{equation}

\noindent In order to keep these expressions short they use a slightly
higher threshold than was used to create Figure \ref{XRef-Figure-82211324}b,
because this suppresses some of the reconstruction vector components
linked to $(x_{1},x_{2},x_{3},x_{4})$. In this particularly simple
example it is possible to obtain very short symbolic expressions,
but more generally continuous-valued computations would be needed
to obtain good approximations to the\ \ network outputs.
\begin{figure}[h]
\begin{center}
\includegraphics{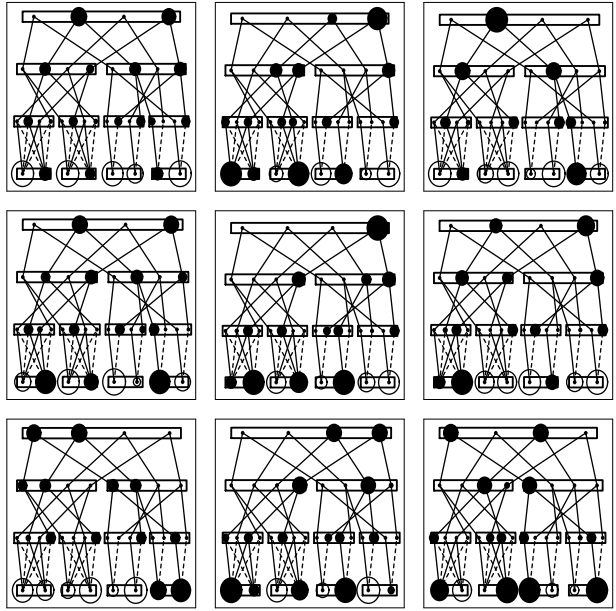}

\end{center}
\caption{Some typical examples of node activities $\Pr ( y_{l}|x_{l-1})
$ (for $l=1,2,3$) in the trained 3-stage network of linked SVQs.
The permuted version of the network is used to make the results
easier to interpret. The area of each filled circle is proportional
to the activity it represents, and the negative values that occur
in the input layer are represented by unfilled circles. The hierarchical
structure of the network is indicated by drawing a box around each
part of each network layer that acts as a separate encoder, so that
typically there are one or two active nodes within every box.}\label{XRef-Figure-822111033}
\end{figure}

Figure \ref{XRef-Figure-822111033} shows some examples of the node
activities $\Pr ( y_{l}|x_{l-1}) $ (for $l=1,2,3$) in the trained
network shown in Figure \ref{XRef-Figure-82211324}c. The individual
pieces of each factorial encoder are indicated by the boxes, and
the patterns of activity are such that every box contains one or
more active nodes, as would be expected if each box were acting
as a separate encoder.
\begin{figure}[h]
\begin{center}
\includegraphics{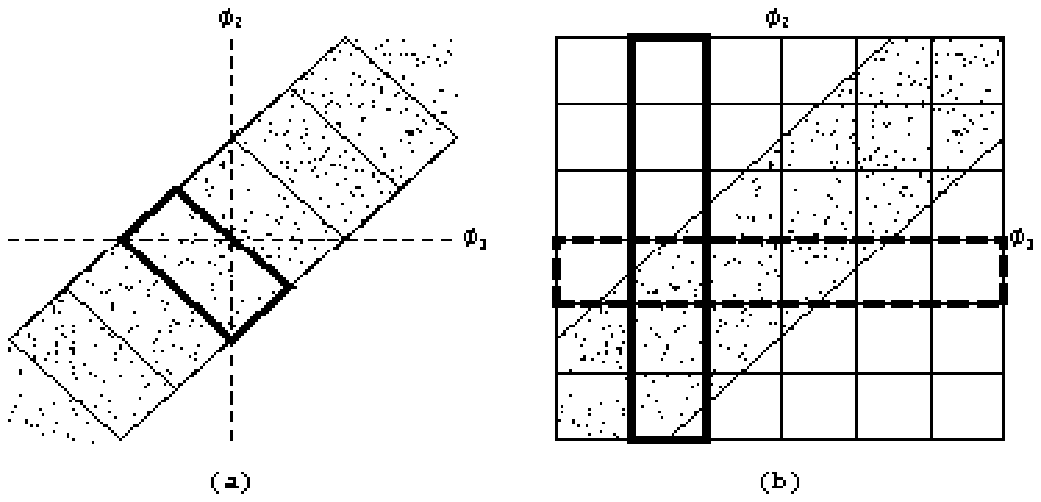}

\end{center}
\caption{ Two ways of encoding a pair of correlated phases $\phi
_{1}$ and $\phi _{2}$. The co-occurrence matrix of $\phi _{1}$ and
$\phi _{2}$ is represented as a narrow band that is populated by
data points, so that the correlation manifests itself as $\phi _{1}\approx
\phi _{2}$. (a) This shows how an invariant encoder operates, in
which the response region of each node is oriented so that it has
high resolution for the $\phi _{1}+\phi _{2}$ but is completely
insensitive to $\phi _{1}-\phi _{2}$. This does not encode information
about the small degree of freedom measured {\itshape across} the
band of the co-occurrence matrix. (b) This shows how a factorial
encoder operates, in which the response region of each node is highly
anisotropic, with high resolution for one of the phases but completely
insensitive to the other phase. Accurate encoding is achieved by
using the nodes in {\itshape pairs} with orthogonally intersecting
response regions, as shown in the example highlighted in the diagram.}\label{XRef-Figure-82211127}
\end{figure}

Figure \ref{XRef-Figure-82211127} shows simplified versions of the
two types of encoder/decoder that occur in Figure \ref{XRef-Figure-82211324}
overlaid on a co-occurrence matrix of the type shown in Figure \ref{XRef-Figure-822105946}.

Figure \ref{XRef-Figure-822111559} and Figure \ref{XRef-Figure-82211165}
show the encoders that occur in stage 1 of Figure \ref{XRef-Figure-82211324}.
These are all factorial encoders of the type shown in Figure \ref{XRef-Figure-82211127}b,
as can be seen from the orientation of the response regions for
the various nodes which cuts across the band of the co-occurrence
matrix in the same way as in Figure \ref{XRef-Figure-82211127}b.
This corresponds to the connectivity seen in Figure \ref{XRef-Figure-82211324}c
where each $\phi _{i}$ has its own encoder.

Figure \ref{XRef-Figure-822111928} and Figure \ref{XRef-Figure-822111935}
show the encoders that occur in stage 2 of Figure \ref{XRef-Figure-82211324}.
These are invariant encoders of the type shown in Figure \ref{XRef-Figure-82211127}a,
as can be seen from the orientation of the response regions for
the various nodes which cuts across the band of the co-occurrence
matrix in the same way as in Figure \ref{XRef-Figure-82211127}a.
This corresponds to the connectivity seen in Figure \ref{XRef-Figure-82211324}c
where each of $\phi _{1}+\phi _{2}$ and $\phi _{3}+\phi _{4}$ has
its own encoder.

Figure \ref{XRef-Figure-82211249} and Figure \ref{XRef-Figure-822112415}
show the encoder that occurs in stage 3 of Figure \ref{XRef-Figure-82211324}.
This is an invariant encoder of the type shown in Figure \ref{XRef-Figure-82211127}a,
which corresponds to the connectivity seen in Figure \ref{XRef-Figure-82211324}c.

The diagrams in this section show how a 3-stage chain of linked
SVQs of the type shown in Figure \ref{XRef-Figure-822105530} self-organises
to process hierarchically correlated phase data. Stage 1 makes an
approximate copy of the data where each phase is separately encoded,
then stage 2 encodes the output of stage 1 discarding the smallest
degrees of freedom, and finally stage 3 encodes the output of stage
2 discarding the next smallest degree of freedom. The chain of SVQs
has thus split itself into a hierarchical network of linked encoders
that is optimally matched to the task of mapping from the original
data at the input to the chain to the compressed representation
at the output of the chain. This is true self-organisation of multiple
encoders unlike the hard-wiring of encoders that is used in other
approaches (e.g. \cite{RossZemel2003}).

\clearpage

\section{Conclusions}

This paper has shown how it is possible to map a data manifold into
a simpler form by progressively discarding small degrees of freedom.
This is the key to self-organising data fusion, where the raw data
is embedded in a very high-dimensional space (e.g. the pixel values
of one or more images), and the requirement is to isolate the important
degrees of freedom which lie on a low-dimensional manifold. A useful
advantage of the approach used in this paper is that it assumes
only that the mapping from manifold to manifold is organised in
a chain-like topology, and that all the other details of the processing
in each stage of the chain are to be learnt by self-organisation.
The types of application for which this approach is well-suited
are ones in which separation of small and large degrees of freedom
is desirable. For instance, separation of targets (small) and jammers
(large) is relatively straightforward using this approach \cite{Luttrell2002}.

Data that is not embedded in a higher-dimensional space is usually
not suitable for processing with the approach used in this paper.
For instance, categorical (or symbolic) data that has one of only
a few possible states is not suitable, but a smoothly variable array
of pixel values is suitable. This type of network is intended to
operate on raw sensor data rather than pre-processed data, and typically
will use high-dimensional intermediate representations in its processing
chain. Because of its ability to compress the raw data into a much
simpler form, this type of network would typically be used as a
bridge between the sub-symbolic raw sensor data and the symbolic
higher level representation of that data.

Although the full connectivity between adjacent layers of the chain
implies that the computations can be expensive, after some initial
training the factorial structure of the various encoders becomes
clear and can be used to prune the connections to keep only the
ones that are actually used (i.e. usually only a small proportion
of the total number). When the chain is fully trained each code
index typically depends on only a small number of contributing inputs
(i.e. a receptive field) from the previous stage of the chain. Furthermore,
because of the normalisation used in each layer, the size and shape
of the receptive fields mutually interact (i.e. there is a fixed
total amount of activity in each layer), so the raw receptive fields
(i.e. as defined by the feed-forward network weights) are {\itshape
different from} the renormalised receptive fields (i.e. after taking
account of normalisation).

The network described in this paper passes information along the
processing chain in a deterministic fashion, because it uses (hypothetical)
histograms containing an {\itshape infinite} number of samples,
which thus do not randomly fluctuate. This was done for computational
convenience (i.e. to avoid Monte Carlo simulations) and is {\itshape
not} a fundamental limitation of the approach used. With additional
computational effort it is possible to operate the network as a
(non-deterministic) Markov chain in which the histograms contain
only a {\itshape finite} number of samples, which therefore randomly
fluctuate and explore network states in the vicinity of the deterministic
state used in this paper.
\section{Acknowledgements}

The research presented in this paper was supported by the United
Kingdom's MoD Corporate Research Programme.

\appendix

\end{document}